\documentclass[sigconf]{acmart}

\usepackage{dirtytalk}

\AtBeginDocument{%
  \providecommand\BibTeX{{%
    \normalfont B\kern-0.5em{\scshape i\kern-0.25em b}\kern-0.8em\TeX}}}
 
\setcopyright{rightsretained}
\acmConference[SIGIR eCom'20]{ACM SIGIR Workshop on eCommerce}{July 30, 2020}{Virtual Event, China}
\acmYear{2020}
\copyrightyear{2020}
\makeatletter
\renewcommand\@formatdoi[1]{\ignorespaces}
\makeatother
\acmISBN{}

\begin{document}

\title[Shopping in the Multiverse]{Shopping in the Multiverse: A Counterfactual Approach to In-Session Attribution}


  
\author{Jacopo Tagliabue}
\authornote{Authors contributed equally to this research project and are listed alphabetically.}
\authornote{Corresponding author.}
\email{jtagliabue@coveo.com}
\affiliation{%
  \institution{Coveo Labs}
  \city{New York}
  \country{USA}
}
  
\author{Bingqing Yu}
\authornotemark[1]
\email{cyu2@coveo.com}
\affiliation{%
  \institution{Coveo}
  \city{Montreal}
  \country{Canada}}

\renewcommand{\shortauthors}{Tagliabue and Yu}

\begin{abstract}
We tackle the challenge of in-session attribution for on-site search engines in eCommerce. We phrase the problem as a causal counterfactual inference, and contrast the approach with rule-based systems from industry settings and prediction models from the multi-touch attribution literature. We approach counterfactuals in analogy with treatments in formal semantics, explicitly modeling possible outcomes through alternative shopper timelines; in particular, we propose to learn a generative browsing model over a target shop, leveraging the latent space induced by \textit{prod2vec} embeddings; we show how natural language queries can be effectively represented in the same space and how ``search intervention'' can be performed to assess causal contribution. Finally, we validate the methodology on a synthetic dataset, mimicking important patterns emerged in customer interviews and qualitative analysis, and we present preliminary findings on an industry dataset from a partnering shop. 
\end{abstract}



\maketitle

\section{Introduction}

\begin{quote}
``Futures not achieved are only branches of the past: dead branches.'' -- Italo Calvino, \textit{The Invisible Cities}.
\end{quote}

Simon searches for ``running shoes'' on the sport apparel eCommerce \textit{Balls\&Things}: he clicks on a pair of shoes, does not love them, goes back to the running section and starts browsing in other categories: socks, t-shirts, headbands... finally, Simon finds a pair of shorts that he really likes, adds them to cart and completes the transaction. Later that day, Alice,~\textit{Balls\&Things} director of digital marketing, is looking at~\textit{Google Analytics} conversion dashboard: Simon's session is not just a win (i.e.~\textit{conversion}), it is a win for~\textit{on-site search}.

~\textit{Is it, though}? In this case, it seems pretty obvious that the search behavior and the conversion event are only mildly related, but some other cases are subtler: did our search engine lead Simon to the purchase, or would he have bought shorts anyway? \textit{On-site attribution} is the task of determining the value of each on-site customer touchpoint, such as on-site search, that leads to a conversion. Addressing the challenge in a principled way is important to many stakeholders: Alice, who needs to allocate budget depending on conversion signals; John,~\textit{Balls\&Things} CTO, who needs to measure search performance; Jean, director of UX, who is redesigning the site experience. In this~\textit{short} paper, we present preliminary results in taking a causal approach to on-site search attribution. In particular:

\begin{itemize}
    \item we phrase the on-site attribution challenge as a \textit{causal} inference: did the search interaction cause the conversion? In turn, we model causality in a counterfactual fashion: would Simon have bought shorts anyway without that interaction? We propose a novel assessment of counterfactual statements in a dense, high-dimensional space, drawing from the notion of \textit{possible world} popularized by modal logic~\cite{Lewis1973-LEWC-2} -- if we could observe a parallel timeline in which Simon does not search for ``running shoes'', we could verify what happens to the purchase event;
    \item we present a deep learning framework allowing for the probabilistic generation of alternative timelines; our solution has minimal barriers to adoption, it requires nothing but essential data tracking and it is straightforward to extend to other functionalities (e.g. recommendation, category listing);
    \item we test the proposed method on synthetic and industry datasets; we show how to simulate ``search interventions'', highlight interesting cases and compare it with industry tools.
\end{itemize}

We believe our findings are interesting to a broad audience: big eCommerce sites with home-grown APIs in need of more accurate measuring tools; mid-size eCommerce sites evaluating external providers and strategic initiatives; multi-tenant SaaS providers that need to measure and communicate the ROI brought by adoption.

\section{Going Beyond Rules: An Industry Perspective}
\label{sec:usecase}
\begin{figure}
  \centering
  \includegraphics[width=\linewidth]{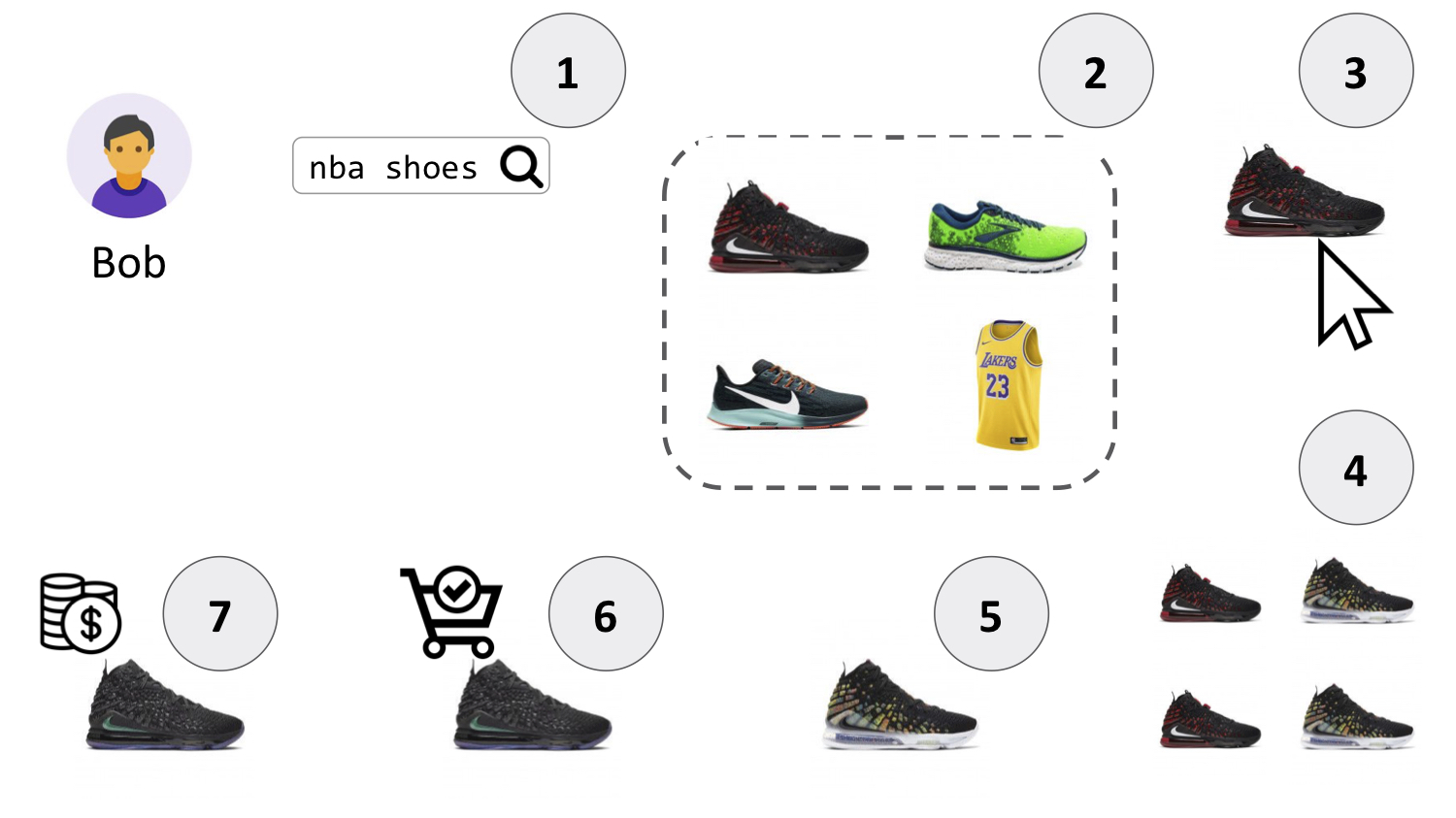}
  \caption{Bob's session on the sport apparel shop: his journey starts with a query (\textit{1}) and a click (\textit{3}) from the SERP (\textit{2}); he then goes up one level to the category page for \textit{basketball shoes} (\textit{4}), browses another pair of shoes (\textit{5}), finally adds the product to the cart (\textit{6}) and buys it (\textit{7}).}
  \Description{A sample shopper journey: was the search interaction responsible for the final conversion event?}
  \label{fig:bobjourney}
\end{figure}

Consider now Bob The Shopper, and his customer journey in Fig.~\ref{fig:bobjourney}: conversion happens after a mix of browsing (\textit{4},~\textit{5}) and search behavior (\textit{1},~\textit{2},~\textit{3}). When Alice looks at Bob's behavior through the lens of industry standard tools -- such as \textit{Google Analytics} (henceforth \textit{GA}) or \textit{Adobe Omniture} - Bob's session features in conversion reports as a search win\footnote{``Conversions are calculated for sessions that include at least one search on your website.'', from \url{https://support.google.com/analytics/answer/1032321?hl=en}.}, as~\textit{GA} just considers whether \textit{any} search interaction was present at some point in a converting session. Since shoppers using search are arguably more motivated to buy in the first place, framing attribution in this way risks conflating search efficiency and prior intent into one measure: if Bob The Shopper searches for ``nba shoes'' and then buys a headband after extensive browsing on the website, how much credit should we attribute to the search engine?

In Bob's case in Fig.~\ref{fig:bobjourney}, it can be argued that search results containing basketball shoes prompted the user to explore first a particular product (\textit{3}), then the general category page (\textit{4}) and finally after some browsing, purchasing basketball shoes (\textit{7}). The intuitive reasoning behind this example is at the heart of the current proposal: i) \textit{attribution} is a causal relation: search interaction is important only \textit{insofar} as it is causally involved in the conversion; ii) \textit{attribution} is not a binary concept, but a matter of degree; iii) in-session \textit{attribution} is crucially linked to shopper intent: if Bob is looking for basketball-related things, and search points Bob - so to speak - in the ``right direction'', we are inclined to attribute the final win to the search engine. The model we propose in Section~\ref{sec:model} is our attempt of making this type of reasoning more precise.

From an industry perspective, it is important to note that calculating causal attribution via continuous A/B testing is not generally feasible, as switching off the search box for a significant portion of users will bring a harmful impact on business revenue; moreover, when \textit{multiple} providers are involved, A/B tests are harder to manage as more players need to be coordinated to assure the validity of the final results. Generally speaking, our findings support an \textit{holistic} view of A.I. services within a retailer. Search contribution to revenues depends on many factors: UX, site structure, number of SKUs, etc. Deciding how much to invest in Information Retrieval technologies should be part of a general strategy of digital growth more than just a local optimization.~\textit{Coveo} is a multitenant SaaS provider with more than 500 clients, ranging from mid-sized business to Fortune 500 companies:~\textit{this} paper is therefore part of a larger project combining research and product development.

\section{Related Work}
\label{sec:relatedWork}
\textit{This} short contribution sits at the intersection of multiple fields: we briefly survey existing literature by main topic.

\begin{figure}[h]
  \centering
  \includegraphics[width=\linewidth]{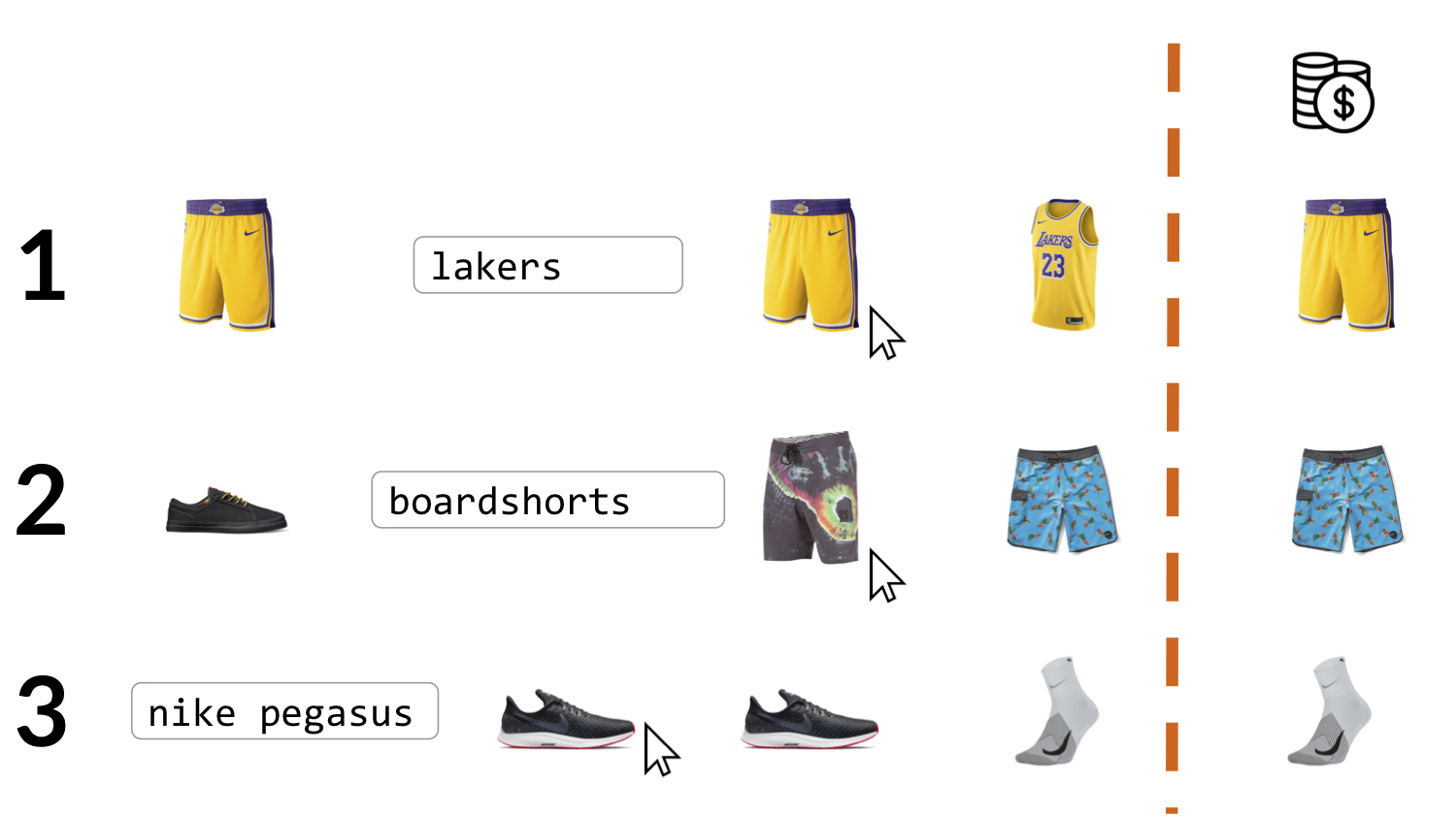}
  \caption{Sample converting sessions from \textit{Shop X}, illustrating  important patterns: \textit{Session 1} is a single-intent session; \textit{Session 2} is a multi-intent session, showing how the search engine effectively allows the shopper to move across the website; \textit{Session 3} shows how search interactions may sometime be successful -- the user clicks on a product -- but the final purchase is unrelated.} 
  \Description{Sample converting sessions from \textit{Shop X}.}
  \label{fig:patterns}
\end{figure}

\textbf{Channel Attribution}. The problem of attributing conversion to known prior events is a well studied problem in marketing~\cite{marketingAds} and, more recently, machine learning~\cite{10.1145/2020408.2020453}; recent work in deep learning on attribution includes LSTM-based frameworks such as ~\cite{Li2018DeepNN,Yang2020InterpretableDL}. While marketing campaigns could be included in our model, we measure in-session attribution focusing on fine-grained user interactions. There is no action taken outside the target website, and no feature collection is required for user identification or segmentation. It should also be stressed that predicting a conversion event \textit{per se} (as in~\cite{Yang2020InterpretableDL}) may fail to detect the subtle differences among interactions occurring within a session, and thus lead to a biased estimation of causal influence; for example, if session intent is very clear (Fig.~\textbf{\ref{fig:patterns}.1}), search interactions may be predictive but still somehow not very \textit{influential}: by taking a counterfactual perspective, we are trying to isolate conversions that would have \textit{not} happened anyway.

\textbf{Browsing Models for eCommerce}. LSTM-based neural networks~\cite{rakuten,Bigon2019PredictionIV} have recently been proven to be effective in modelling long-range dependencies in eCommerce browsing. Dense representations for products have been popularized by~\textit{prod2vec} models~\cite{Grbovic15, Vasile16}: to the best of our knowledge, \textit{this} is the first work analyzing in-session trajectories through a unique representational space, which includes product embeddings and possibly infinite linguistic behavior.

\textbf{Causal Inference}. Inferring causal relations from observational data is a well-studied topic in philosophy~\cite{10.5555/1756006.1859905}, econometrics~\cite{ECO-014} and classical machine learning~\cite{10.5555/331969}. In an eCommmerce setting,~\cite{Sharma2015EstimatingTC} estimates the causal importance of recommendations by exploiting instantaneous shock in traffic: data constraints, both in times -- nine months -- and in traffic -- 2M unique users, challenge the widespread applicability of the method to other shops. Matching methods are another common substitutes for randomized trials~\cite{stuart2010}, especially in the medical literature: however, finding two shoppers with, say, $5$ matching interactions in a shop with 25k products is almost impossible -- in other words, while  some of the ideas below are philosophically similar to matching methods, traditional approaches are hard to scale to the dimensions of our space. In the deep learning tradition,~\cite{Nauta_2019} proposes a CNN-based technique for multivariate time-series modelling; however, their interest is mostly in understanding how randomly changing a real number affects predictability, which is a much narrow counterfactual than the ones needed for our use case -- our proposal crucially relies on taking seriously the multi-dimensional space of possibilities in front of a shopper that lands on a target shop. Standard packages aimed at democratizing causal reasoning~\cite{dowhy} are geared towards traditional statistical methods (regression, random forests etc.) and they cannot scale easily to handle the high-dimensional space of products and linguistic interactions. Finally, for (mostly) nostalgic reasons, it is worth remembering that linking causation to counterfactuals was first done by David Hume\footnote{More precisely, in \textit{An Enquiry Concerning Human Understanding}, Section VII.} and further improved by David Lewis~\cite{10.2307/2025310}: our method owes a great debt to their revolutionary ideas.

\section{Methodology}
\label{sec:methodology}
Since gold labels of~``true attributions'' are not available, we proceed to validate our methodology in two steps: first, we analyze hundreds of converting search sessions from partnering~\textit{Shop X}, extract important patterns and produce a synthetic dataset with known proportions of ``causally relevant interactions''. Second, once we confirm that our methodology successfully detects interesting patterns, we apply it to \textit{Shop X} test set and compare the results with other attribution strategies. 

\subsection{The ``Attribution Zoo''}
\label{sec:zoo}

We sampled and analyzed 500+ converting sessions from \textit{Shop X} (Section~\ref{sec:dataset}) \textit{including search interactions} (query issued by user is followed by one or more products clicked by the shopper). We classify major patterns according to two independent dimensions:

\begin{enumerate}
    \item \textit{relevance of interaction for purchase}: some interactions are~\textit{directly} linked to a purchase, as the shopper buys exactly the product found through search (Fig.~\textbf{\ref{fig:patterns}.1}); on the other side of the spectrum, some interactions are completely irrelevant: a shopper searches for ``sneakers'' but then buys socks (Fig.~\textbf{\ref{fig:patterns}.3}). In the middle, cases like Fig.~\ref{fig:bobjourney} -- search interactions is semantically related to the purchase, but product is not exactly the same;
    \item \textit{breadth of shopping intent}: most sessions are single-intent sessions, such that browsing, searching and buying happens within a somewhat specific category of products (e.g. \textit{sneakers}, \textit{lakers merchandise}, etc. -- Fig.~\textbf{\ref{fig:patterns}.1}). Other sessions are \textit{multi}-intent: the shopper starts with an intent, e.g. sneakers, and then uses the search box to quickly move to another intent, swimsuit (Fig.~\textbf{\ref{fig:patterns}.2}). Since our model is based on the product space induced by user browsing (Section~\ref{sec:prod2vec}), a broader range of intent corresponds to the physical equivalent of shoppers visiting several aisles of a shop. In this analogy, search interactions are a very handy way of ``travelling'' a great distance with a single action.
\end{enumerate}

The result of the extensive qualitative analysis was validated through interviews with Customer Success Managers and target clients, confirming our intuitions: \textit{ceteris paribus}, more \textit{relevant} interactions should be considered more causally influential for conversion; \textit{ceteris paribus}, the search interaction in Fig.~\textbf{\ref{fig:patterns}.2} has greater influence on conversion than Fig.~\textbf{\ref{fig:patterns}.1}, since in Fig.~\textbf{\ref{fig:patterns}.2} the search engine is instrumental in answering the shopper's inquiry, moving the shopper to a very different region of the product space. These important observations and customer insights are used to build a synthetic dataset out of real shopping events.

\subsection{Datasets}
\label{sec:dataset}
Our first dataset is a synthetic dataset (\textbf{SD}) built by composing sub-sessions from the available data on \textit{Shop X} -- please see Appendix~\ref{app:synthetic} for the details.~\textbf{SD} has a total of $1,075,000$ sessions, $375,000$ with search; sessions with search reflect the main patterns from Section~\ref{sec:zoo} in known proportions ($175,000$ relevant, $200,000$ irrelevant); single-intent vs multi-intent sessions are generated with specific code paths, to highlight the qualitative differences within the \textit{relevant} bucket. Our second dataset is an industry dataset (\textbf{ID}), obtained by collecting anonymized sessions from June 2019 to September 2019 from our partnering \textit{Shop X} -- we use $912,884$ sessions for \textit{prod2vec} training (Section~\ref{sec:model}) and we then run our benchmarks on a total of 10K sampled converting sessions with search.~\textit{Shop X} is a mid-size eCommerce, has 26k distinct SKUs, with Alexa Ranking >150k.~\textit{Shop X} was an ideal first candidate for this project, since it doesn't leverage recommendation or listing APIs, but only an external search provider.

\section{The Shopper Multiverse}
\label{sec:model}

The main intuition behind our model is that Bob's session (Fig.~\ref{fig:bobjourney}) is best conceptualized as a path in a vector space, more than a Markov-like sequence of actions~\cite{doi:10.1137/1.9781611972733.10}. At a first approximation, we can describe our strategy for calculating attribution as two-fold: first, we build a \textit{shared} vector space in which both product browsing and search interactions can live -- if events are points in a multi-dimensional space, converting sessions are specific \textit{paths} in that space; second, leveraging the full differentiability of shopping paths, we train a deep recurrent neural network over historical sessions to build a simulation model of browsing: when evaluating a session, we run the model with selected perturbations to assess how much the outcome (i.e. the conversion) would change if events (i.e. the interactions with search) had been different. We now proceed to specify the technical components in detail.

\subsection{Building a shared vector space}
\label{sec:prod2vec}

We prepare dense vectors for all the products in the target shop. Product embeddings are trained with the CBOW negative sampling~\cite{Mikolov:2013,Mikolov2013EfficientEO}, by swapping the concept of words in a sentence with products in a browsing session (from \textit{word2vec} to \textit{prod2vec})~\cite{Grbovic15}; ETL and hyperparameter optimization follows the guidelines presented in~\cite{AuthorsKDD2020}; for this task, we use interaction-specific embeddings~\cite{ETSYECNLP20}, such that a product with $SKU=xyz$ is embedded in the space as $xyz_{detail}$, $xyz_{click}$, $xyz_{purchase}$. Given a product $p$ and interaction $I$, we denote the associated embedding as $V(p_I)$. Starting from this ``shop space'', we use the same intuition behind~\textit{Search2Prod2Vec}~\cite{CoveoECNLP22} as our query embedding strategy: for a query $q$, we take the top $N$ items returned by the engine, $p1_{detail}$, $p2_{detail}$, $... pn_{detail}$ and then create a \textit{deep set}~\cite{Slch2019OnDS} by computing the average of $V(p1_{detail})$, $V(p2_{detail})$, $... V(pn_{detail})$; in other words, we take the engine response as the~\textit{denotation}~\cite{Tagliabue2019LexicalLA} of the issued query, so that the~\textit{meaning} of $q$ is a function from $q$ to a set of products falling under that concept~\cite{10.5555/335289}\footnote{Empirically, we determined $N=20$ to be a good threshold.}.

\subsection{Training a generative browsing model}
\label{sec:trainlstm}

 \textit{LSTM}s have obtained SOTA results in language~\cite{Melis2018OnTS} and browsing~\cite{Bigon2019PredictionIV} modelling. In our use case, we exploit both the sequential and generative~\cite{10.5555/3104482.3104610} nature of \textit{LSTM}s: in particular, since the trained network models the conditional probability of a product given the previous ones, $P(x_n|x_{0},... x_{n-1})$, we will use this distribution to guide interventions in a principled way (as opposed to pure random permutation~\cite{Nauta_2019}). Our model is trained using~\textit{Shop X}'s usage logs, by feeding 
 our pre-trained \textit{prod2vec} embeddings as input at each timestep. For training, we use teacher forcing to pass the vector of the current timestep's target, offset by one position, as the input for the next timestep~\cite{Williams89alearning}. Hence, once trained, the model can start with any given input sequence and use autoregression sequence generation to predict the tokens for the next timesteps~\cite{Bahdanau2015NeuralMT}. In addition to human inspection, we also obtain a $HR@1=0.16$ over a hold-out dataset, which confirms LSTM ability to capture user behaviour, as compared to similar models in the product embeddings literature~\cite{Vasile16}. In order to use the LSTM model to generate our alternative timelines, we experiment with different sampling methods that have proven to work well for non-deterministic sequence generation: \textit{Top-K sampling}~\cite{Fan2018HierarchicalNS}, \textit{Top-K sampling with temperature} and \textit{Top-P sampling}~\cite{Holtzman2020TheCC}. Since all these decoding methods are non-deterministic, we generate $T$ samples per alternative timeline (Section~\ref{sec:attribution}), to reduce the effect of random fluctuations. Preliminary experiments prove that simulation results were consistent across different methods, therefore we only report results obtained with optimal settings: \textit{Top-K} sampling, with $K=3$ and $T=100$.

When fully trained, our browsing model implicitly captures two important dimensions: first, how latent user intent shapes the unfolding of a shopping session; second, how site structure implicitly constraints what product is reachable from what (i.e. two pairs of sneakers may be only two clicks away, sneakers and a basketball four). It is the combination of these two latent properties - what the shopper wants and how easy is to get it just by browsing - that makes the model so appealing for counterfactual reasoning.

\subsection{Assessing causal influence} 
\label{sec:attribution}

$C$ causing an event $E$ is understood in a counterfactual framework as ``$E$ would not have happened, if $C$ had not been the case''~\cite{10.2307/2025310}; in turn, counterfactuals in formal semantics and modal logic are understood through the notion of \textit{possible world}~\cite{Lewis1973-LEWC-2}, alternative realities in which events are different: what we need to find is a possible world $w_1$, which is exactly like the actual world $w_@$ up until $C$ occurred, and verify if $E$ happens there without $C$. Since outer worlds are hard to come by, counterfactual causation is often framed through~\textit{interventions}~\cite{Woodward2003-WOOMTH}, i.e. hypothetical changes made to an upstream variable in a causal chain to verify the downstream effect on the target. In Fig.~\textbf{\ref{fig:patterns}.2}, an \textit{intervention} would be to redirect the shopper to the sneakers category after the first item interaction, instead of letting the shopper search for ``boardshorts'': a key appeal of the model is its geometrical intuitive interpretation -- the more the timeline \textit{post-intervention} ends up far in the shop space from the original purchase, the more the target interaction causally explains the outcome.

\begin{figure}
  \centering
  \includegraphics[width=\linewidth]{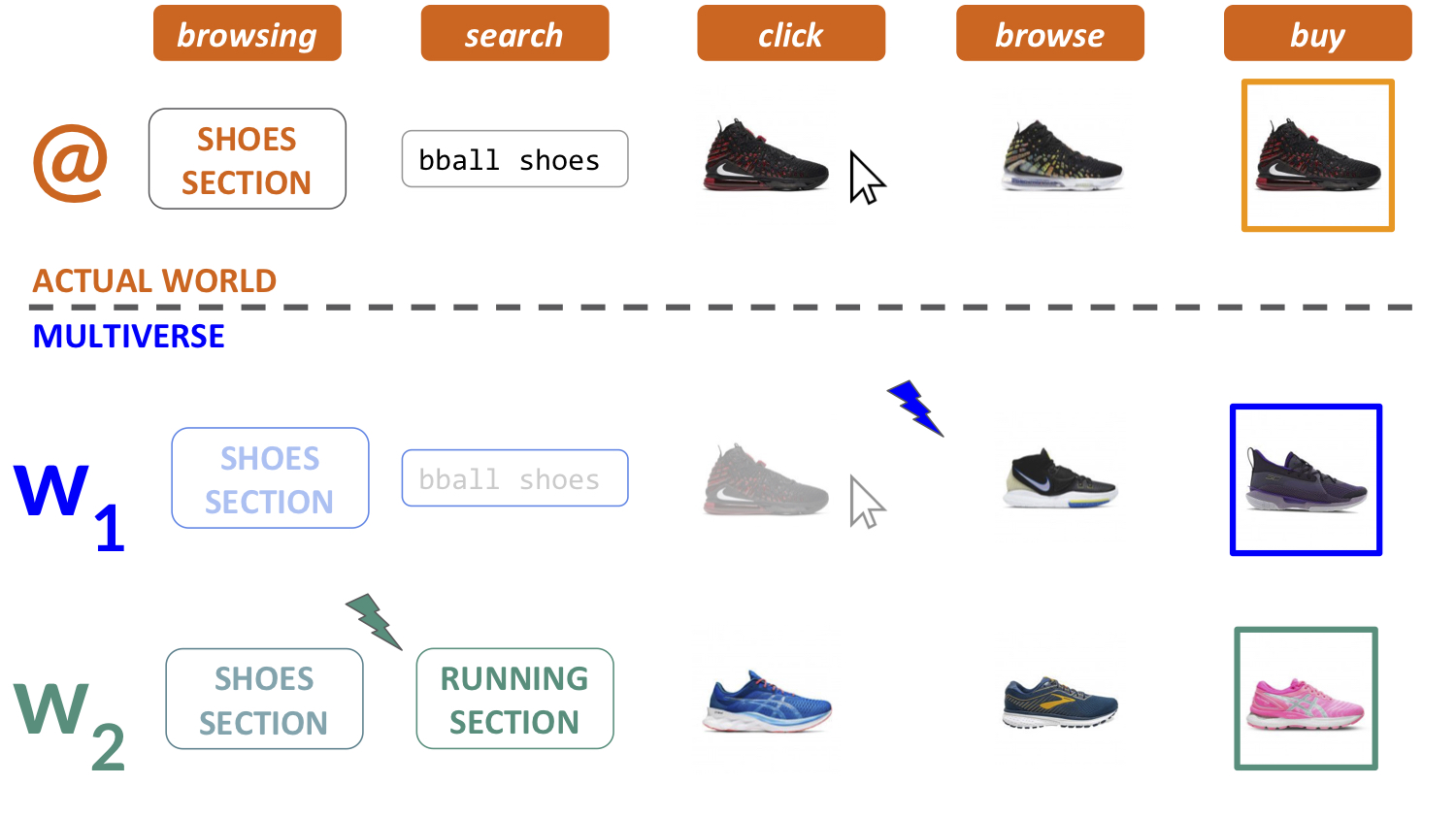}
  \caption{Causal influence is a function of three timelines in the shop space: the actual converting session ($@$, \textit{orange}); a \textit{blue} session (for simplicity, shown here as a single session and not an average across many simulations), where the future diverges \textit{after} the interaction; a \textit{green session}, where the future diverges immediately \textit{before} the query is issued.}
  \Description{Shopper journeys are paths in the shop space.}
  \label{fig:intervention}
\end{figure}

More precisely, to measure the change brought by our intervention, we take inspiration from the \textit{difference in difference} approach (\textit{DD}) for time-series, popular in econometrics~\cite{RePEc:now:fnteco:0800000014,RePEc:nbr:nberwo:8841}. In particular, the causal influence of the search interaction is a function of the relative position in the product space of the ending spot for three timelines (Fig.~\ref{fig:intervention}): i) the original converting session including a search interaction ($@$); ii) an alternative timeline, generated by keeping constant all events until \textit{after} the interaction (click on product), and then letting the model unfold into the future ($w_1$); iii) an alternative timeline, generated by keeping constant all events \textit{before} the query, and then letting the model unfold into the future ($w_2$). Since shoppers' paths live in a dense product space, an easy way to quantify the \textit{DD} is by making use of the cosine distance in the embedding space. After timeline generation, we obtain two cosine distances by comparing the original purchase spot to the ending spots of our two alternative timelines. To constrain the length of generated samples and allow faster computation, we let the decoding process to stop once the original purchase timestep is reached. As mentioned in Section~\ref{sec:trainlstm}, we use \textit{Top-K} sampling to generate 100 samples for each type of alternative timelines, and take the average of their ending spots to represent the ending points for $w_1$ and $w_2$. It is the relative ending position of $@$, $w_1$ and $w_2$ that captures the counterfactual importance of search -- for Fig.~\textbf{\ref{fig:patterns}.1}, all paths will end in a similar region, for Fig.~\textbf{\ref{fig:patterns}.2}, $@$ and $w_1$ will be closer, with $w_2$ far apart. Several examples of actual and generated paths are collected in Appendix~\ref{app:examples}.

\section{Experiments}
\label{sec:discussion}
In the absence of golden labels, we first run the proposed multiverse method (\textbf{MV}) on our synthetic dataset \textbf{SD}, and then run it on actual data comparing the computed attribution with other industry methods.

\subsection{Results on SD}
\label{sec:SD_res}
We run \textbf{MV} on \textbf{SD} as a double sanity check: \textit{quantitatively}, we are looking at evidence that the proposed method is able to distinguish relevant vs irrelevant search interactions; \textit{qualitatively}, we are looking at evidence that the proposed method is capturing counterfactual nuances we care about, e.g. the difference between  Fig.~\textbf{\ref{fig:patterns}.1} and  Fig.~\textbf{\ref{fig:patterns}.2}. To obtain an accuracy measure, we first considered writing a decision module with manual rules that would encode the geometrical intuitions of Appendix~\ref{app:examples}; however, it turned out to be more practical to train a small multilayer perceptron to learn the decision boundary starting from the three timelines, instead of coming up with hard-coded rules. After training the binary classifier on 80\% of \textbf{SD}, we test \textbf{MV} using hold-out samples and compare predictions with the synthetic labels, obtaining as accuracy $0.93$, recall $0.93$ and F1 $0.93$. Outside of the binary context, we also report a ``geometrical interpretation'' of the timelines generated by \textbf{MV}, for the four types of causal patterns to be found in \textbf{SD}. The \textit{Distance Score} (\textit{DS}) is calculated starting from the cosine distance (\textit{d}) between the last event in $w_1$ and $@$, and between $w_2$ and $@$: $DS=d(w_2,@) / d(w_1,@) * (1 / d(w_1,@))$; in other words, \textit{DS} captures how far from the actual events the generated timelines are (using $d(w_1,@)$ as a normalizing factor). The \textit{DS} average values -- after re-scaling and log-normalization -- reported in Table~\ref{tab:finegrained} show from a non-binary perspective \textbf{MV}'s ability to not just distinguish relevant vs irrelevant interactions (positive vs negative \textit{DS}), but also to make subtler distinctions and assign appropriate causal weights. We perform error analysis on misclassified samples and identified three patterns for future improvement:

\begin{enumerate}
    \item \textit{low quality embeddings}: when SKUs and/or issued queries suffer from data sparsity, timelines become more erratic; several options for ``cold-start'' embeddings exist in the \textit{prod2vec} literature, with varying degrees of engineering disruption~\cite{Vasile16}, and it is certainly an important insight for future work;
    \item \textit{regression to the mean}: when a somewhat irrelevant search is issued in a session happening in the most crowded region of the space -- i.e. \textit{shoes} --, all timelines still converge, mimicking a relevant pattern;
    \item \textit{sensitivity to pre-search intent}: when many browsing interactions happen before a \textit{relevant} search, \textbf{MV} tends to mark search as irrelevant; while technically those queries were created to be causally connected to the purchase, this behavior is a very interesting side-effect of embracing counterfactual reasoning. In particular, we verified that shortening the sequence of events before search makes \textbf{MV} predict the correct (\textit{relevant}) label: the more shopping intent is displayed before the query, the more the model thinks conversion would have happened \textit{anyway}.
\end{enumerate}

\begin{figure}
  \centering
  \includegraphics[width=\linewidth]{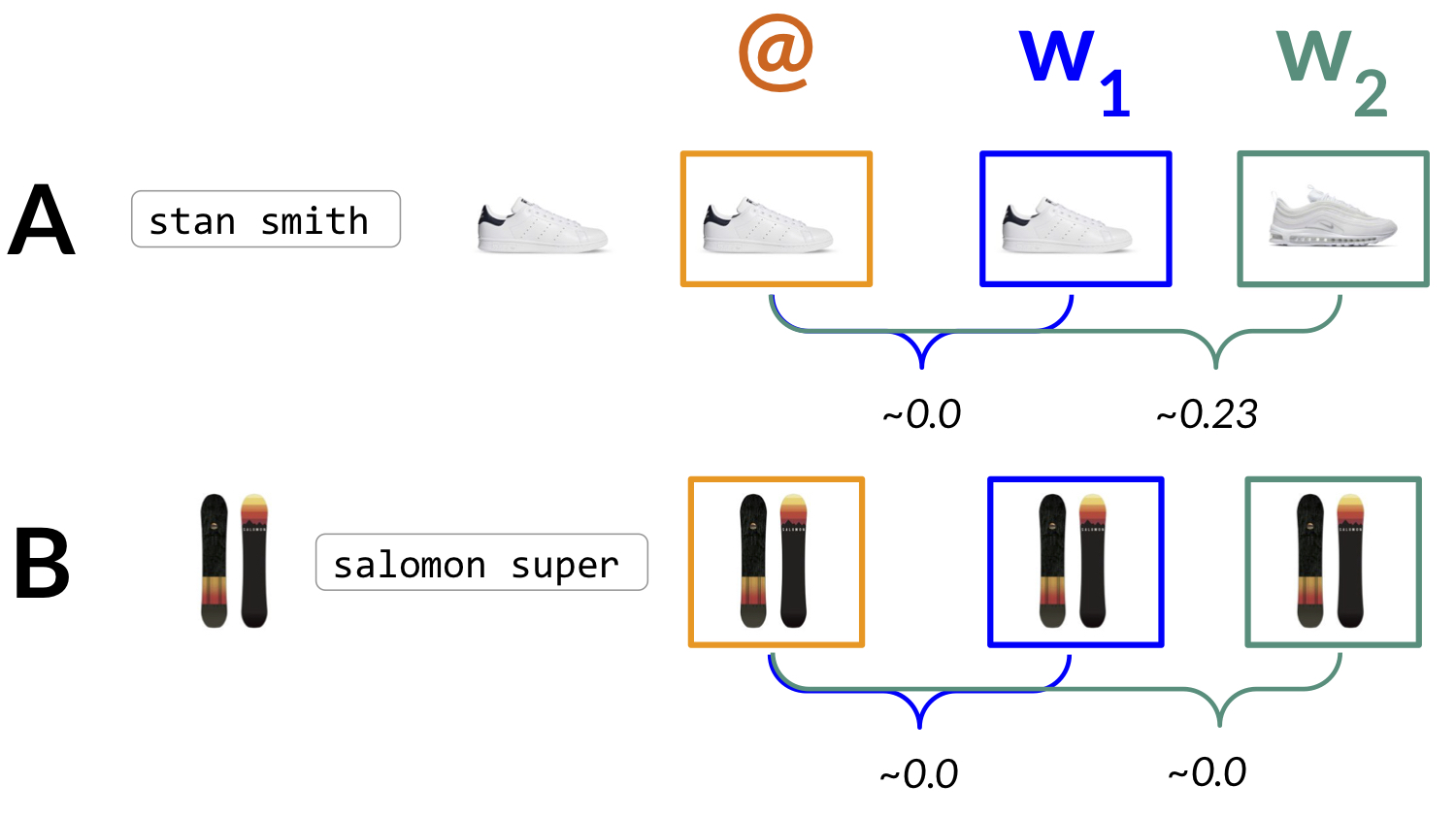}
  \caption{Two converting sessions with search attribution. Session \textbf{A} starts with a query and ends up with purchasing a pair of shoes (\textit{orange}) -- products from alternative timelines (average) are shown in \textit{blue} and \textit{green}, with cosine distance below; session \textbf{B} starts with product browsing and ends in a snowboard purchase after a search query -- the alternative timelines (on average) all converge to the same product.} 
  \Description{Sample converting sessions from \textit{Shop X}.}
  \label{fig:qualitative}
\end{figure}

From a qualitative perspective, we manually inspect sample sessions belonging to different patterns. In particular, after quantitatively validating the overall accuracy of \textbf{MV}, we are now interested to see how counterfactual inference is represented in the model. Consider \textbf{A} and \textbf{B} in Fig.~\ref{fig:qualitative}: they are both converting sessions accurately attributed to the search interaction. However, search plays a different role in them: in \textbf{A}, the shopper lands on the website and search is causally responsible to guide her immediately to fulfill her intent; in \textbf{B}, the shopper is browsing Salomon boards even before issuing a query.~\textbf{MV} is able to capture the difference between those two cases thanks to the $w_2$ timeline: in \textbf{B}, search is less important as the intent is clear enough to make us believe that the shopper would have purchased anyway that board. Please refer to Appendix~\ref{app:examples} for the full charts of the patterns in Fig.~\ref{fig:patterns}. 

\begin{table}
  \caption{Distance Score for four distinct causal patterns: search unrelated to conversion (\textit{CU}), single intent with exact match (\textit{SE}), multi-intent (\textit{ME}), single intent with related-but-not-exact match (\textit{SR}).}
  \label{tab:finegrained}
  \begin{tabular}{lc}
    \toprule
    Pattern&Distance Score\\
    \midrule
    \textit{CU} & -1.062\\
    \textit{SE} & 2.315\\
    \textit{ME} & 3.359\\
    \textit{SR} & 2.262\\
  \bottomrule
\end{tabular}
\end{table}

\subsection{Results on ID}
\label{sec:ID_res}
Experimental results are presented in Table~\ref{tab:results}, reported as percentage of conversions for sessions with search that are actually \textit{attributed} to search interactions by the selected methodology. First, we compare \textbf{MV} with industry standard heuristics, as customary in the multi-touch attribution literature~\cite{Yang2020InterpretableDL}; \textit{GA}-style attribution is a natural upper bound as \textit{any} converting session with search will be counted positively and it is therefore likely to over-estimate the causal impact. Second, we report another common heuristics, a ``click-then-buy'' rule (\textbf{CB}) to the effect that products count as causally related only when purchased items are among after-search clicks; third, we implement a ``semantic similarity'' method (\textbf{SS}), which measures the distance in the embedding space between the target interaction and the purchased product. To simplify the comparison with industry standards, for \textbf{MV} and \textbf{SS} we report the results of the binary classifier, trained for both as described in Section~\ref{sec:SD_res}.

\begin{figure}
  \centering
  \includegraphics[width=\linewidth]{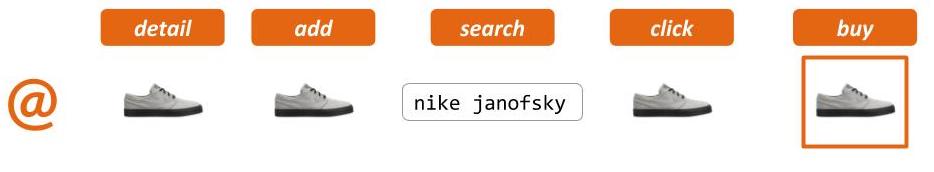}
  \caption{A converting session, featuring browsing and add-to-cart event \textit{before} any query is issued: for \textbf{SS} the query is predictive of the purchase; \textbf{MS}, on the other hand, thinks the shopper was going to buy the shoes \textit{anyway}.} 
  \Description{Sample converting session from \textit{Shop X}, showing how \textbf{MS} and \textbf{SS} may diverge in their interpretation of shoppers.}
  \label{fig:msss}
\end{figure}

We sample interesting sessions from the test set (i.e. sessions where models disagree on the causal interpretation) and make three main observations:

\begin{enumerate}
    \item all \textit{non-GA} models reach the same conclusion as~\cite{Sharma2015EstimatingTC}: once industry heuristics are replaced by stricter rules or more sophisticated inference, the magnitude of direct attribution is less than what was naively thought; it is interesting to note that search impact seems to be significantly bigger than what the literature estimates for recommendations;
    \item \textbf{MV} and \textbf{SS} are more flexible than \textbf{CB}, which relies on a deterministic match that excessively penalizes interactions that are still valuable, such as Bob's query in Fig.~\ref{fig:bobjourney};
    \item \textbf{MV} and \textbf{SS} tend to agree on the overall score, but \textbf{SS} ignores entirely the pre-search intention, and therefore it is behaving as a \textit{prediction} method (``how closely related are interaction and purchase?''), more than a \textit{causal} one (``would Bob have bought those shoes anyway?''). Fig.~\ref{fig:msss} shows a session where predictions from \textbf{SS} and \textbf{MV} diverge: since the shopper issued a query \textit{after} browsing and adding to cart the target product, \textbf{MV} marks the search interaction as weakly relevant, since most alternative timelines would end in the same way. For the reasons explained at length in Section~\ref{sec:usecase}, we believe that the ability to capture this type of causal dynamics is what makes \textbf{MV} so appealing when discussing in-session attribution. 
\end{enumerate}

\begin{table}
  \caption{Search attribution on \textbf{ID} for all the methods.}
  \label{tab:results}
  \begin{tabular}{lc}
    \toprule
    Method&\% Search Attr.~\textbf{ID}\\
    \midrule
    \textit{GA} & 100\\
    \textit{CB} & 47\\
    \textit{SS} & 77\\
    \textit{MV} & 75\\
  \bottomrule
\end{tabular}
\end{table}

Based on the quantitative performance on \textbf{SD}, and the qualitative analysis on both \textbf{SD} and \textbf{ID}, we conclude that we have strong \textit{prima facie} reasons to consider attribution judgments by \textbf{MV} an accurate representation of the underlying causal dynamics. As we stress in the ensuing section, extending the validation to other shops / verticals / services is a natural next step to confirm the generalizability of these findings.

\section{Conclusions and Future Work}
\label{sec:conclusions}
In \textit{this} paper we presented preliminary results towards a more sophisticated understanding of A.I. services in the context of attribution; by leveraging the link between causality and counterfactual reasoning, on one hand, and counterfactuals and generative models, on the other, we were able to frame in-session attribution as a question of model behavior under perturbation -- contrary to approaches to user browsing (and attribution) where state-space is constrained by artificially coarse-grained models, we did it by taking seriously the high-dimensional nature of the underlying space, in which thousands of products and \textit{unbounded} linguistic interactions co-exist. The proposed method, while now applied to on-site attribution for eCommerce search, is general enough to be applicable to any situation in which a generative model can be successfully trained to represent the space of possible actions for the target user. Our findings support the pre-theoretical intuition that~\textit{relevance} with respect to the underlying intent is crucial for conversion. 

While results are very encouraging, our claims are still limited in scope, as we extensively validated only one shop. Our roadmap starts with adding use cases: \textit{Shop X} was an ideal first candidate as we could test search interactions in isolation, but \textit{recommendation}, \textit{category listing} and even marketing actions are straightforward extensions once the multiverse is generalized to multiple services per session. It is also important to highlight that \textit{in-session} attribution, by design, treats all incoming users as \textit{distinct}: no long-term pattern is considered, and all shoppers are treated in the same way; this simplification was justified by \textit{Shop X} statistics -- <9\% of users visited the site for more than 2 times in a year --, but more nuanced treatments are possible. 

It should be obvious that our counterfactuals are reliable only insofar as the timeline generating model is; there may be better generative models for specific cases, such as, for example, sequence-based GANs~\cite{Yu2017SeqGANSG}; furthermore, it should be possible to engineer a ``human-in-the-loop'' evaluation to collect relevance judgments to better align model objectives with human intuitions -- going from a purely binary concept of attribution to a continuous one is in itself an interesting product / UX challenge. Finally, while \textit{do}-calculus~\cite{10.5555/331969} has been historically developed for a discrete, low-dimensional world, it would be interesting to draw explicit and formal connections between our geometric model in high-dimensional spaces and the basic theory of causal inference.

While we do recognize that controlling for all the causal effects in our method is more challenging~\cite{NBERw4956} than in graphical models, we also value tangible \textit{practical progress}: having a fully unbiased estimate of search importance \textit{may} not be possible at scale, but we can (and should) still strive to improve our attribution strategy over heuristics we know \textit{for sure} to be partial.

\begin{acks}
Thanks to Andrea Polonioli and Federico Bianchi for the usual thoughtful comments, and to our reviewers for greatly helping improving the paper. We had great discussions with Ciro Greco, Giovanni Cassani, Eric  Savoie and Charles Fortier on previous versions of this work, and we would like to thank Mattia Pavoni for gathering valuable client feedback. Finally, thanks to Bingqing's mother for giving us a much needed shopper's perspective -- and for great dumplings.
\end{acks}

\bibliographystyle{ACM-Reference-Format}
\bibliography{sample-base}

\appendix

\section{Visualization of Sample Sessions}
\label{app:examples}

\begin{figure}[h]
  \centering
  \includegraphics[width=\linewidth]{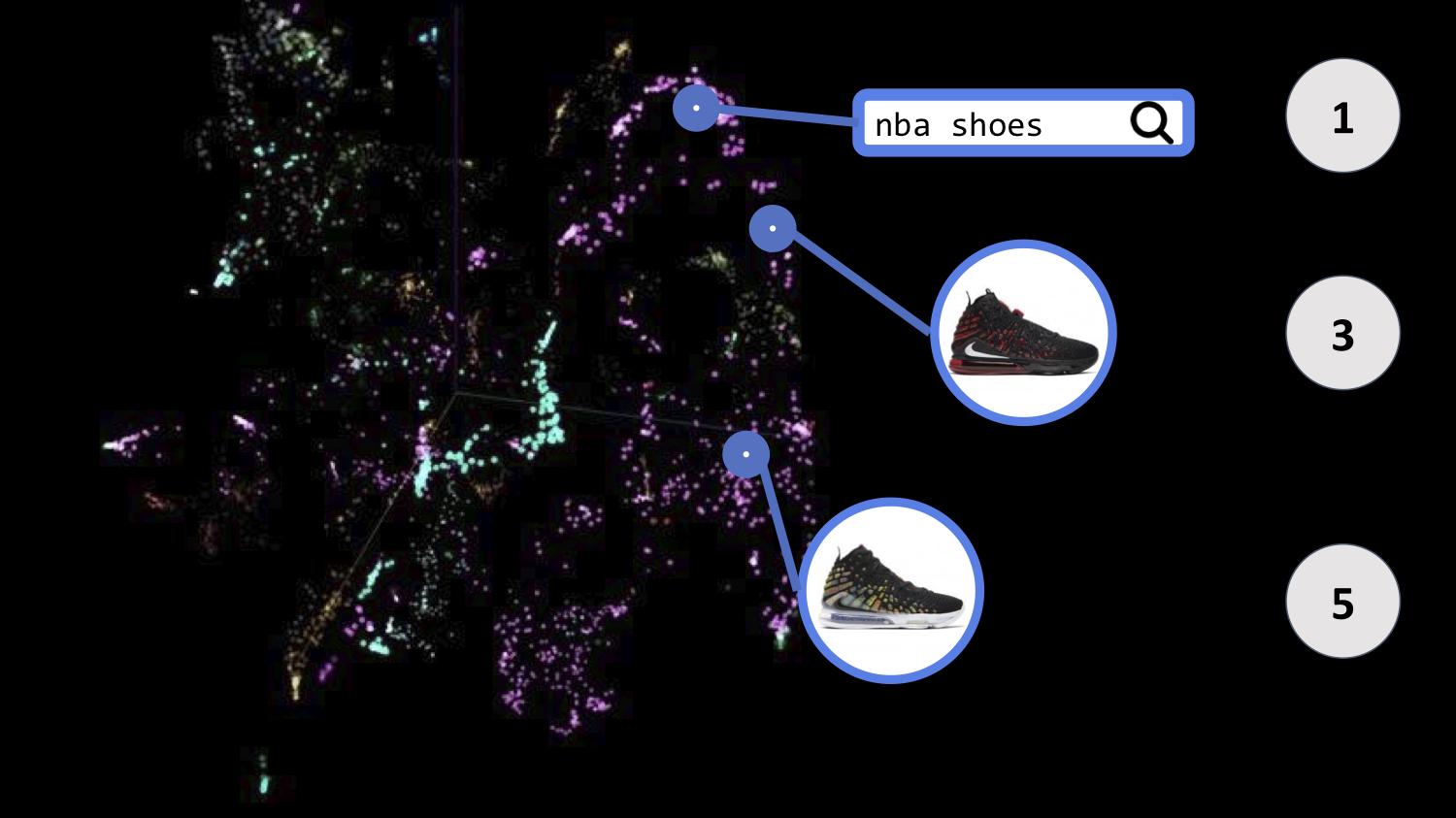}
  \caption{Bob's journey (Fig.~\ref{fig:bobjourney}) in the shop space: by building a shared vector space for browsing interactions and services (search/recommendation), we can model conversions as paths in a dense space. In the image, \textit{prod2vec} embeddings from the target shop are represented through T-SNE; products are color-coded by sport activity (basketball, running, soccer etc.).}
  \Description{Shopper journeys are paths in the shop space.}
  \label{fig:productspace}
\end{figure}

Embedding product interactions and linguistic behavior in a unified dense space (Section~\ref{sec:prod2vec}) allows quite literally to model shopping sessions as \textit{paths} in the underlying space, in which products and queries that are semantically related live close by (Fig.~\ref{fig:productspace}).

\begin{figure}
  \centering
  \includegraphics[width=\linewidth]{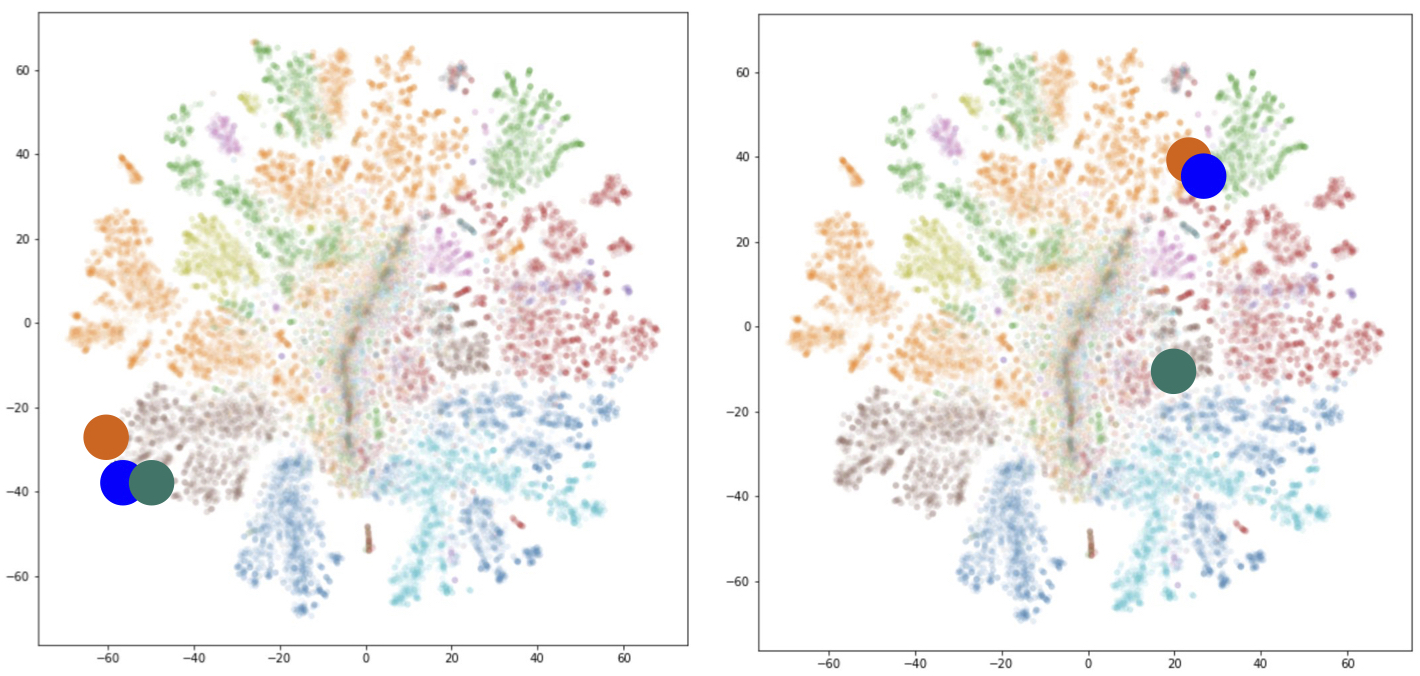}
  \caption{Single-intent (\textit{left}) and multi-intent (\textit{right}) sessions in which the search interaction is causally relevant for the final purchase. The picture shows the ending states of three timelines for the cases in Fig.~\textbf{\ref{fig:patterns}.1} and Fig.~\textbf{\ref{fig:patterns}.2}: \textit{orange} is the actual world (@ in Fig.~\ref{fig:intervention}), \textit{blue} is the timeline generated after the search interaction ($w_1$), \textit{green} is the timeline generated before the search interaction ($w_2$).}
  \Description{Shopper journeys are paths in the shop space.}
  \label{fig:relevant}
\end{figure}

\begin{figure}
  \centering
  \includegraphics[width=\linewidth]{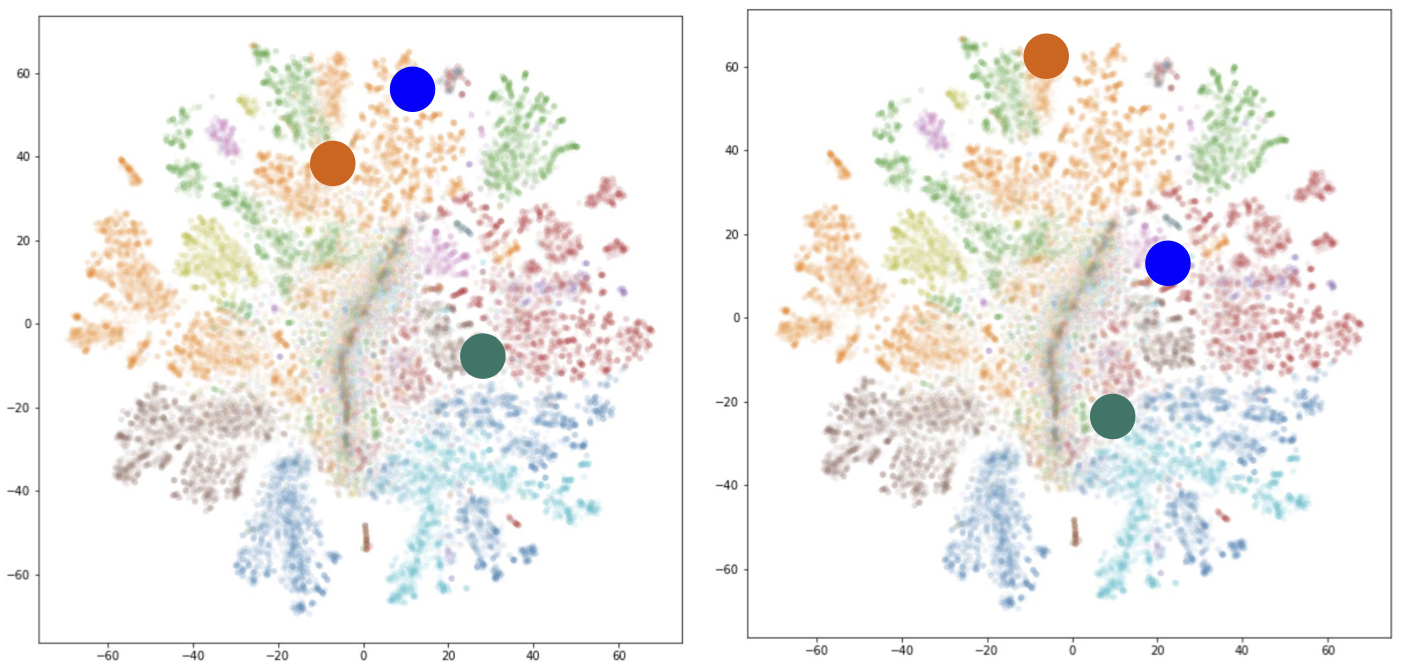}
  \caption{Single-intent (\textit{left}) and multi-intent (\textit{right}) sessions in which the search interaction is \textit{not} causally relevant for the final purchase. The picture shows the ending states of three timelines for the cases in Fig.~\textbf{\ref{fig:patterns}.3} and an additional case, in which search is not relevant and the shopper travels extensively in the underlying space: \textit{orange} is the actual world (@ in Fig.~\ref{fig:intervention}), \textit{blue} is the timeline generated after the search interaction ($w_1$), \textit{green} is the timeline generated before the search interaction ($w_2$).}
  \Description{Shopper journeys are paths in the shop space.}
  \label{fig:irrelevant}
\end{figure}

The attribution model put forward in Section~\ref{sec:attribution} is based on the relative position of the last item in three timelines: the actual converting session we are evaluating, and two sessions generated by simulation through the trained deep neural network. One of the greatest strength of the counterfactual approach we propose is its intuitive representation, and the ability to make principled distinctions between different types of search influence (great, mild, almost nonexistent). We collect here four projections in the product space of representative timelines:

\begin{enumerate}
    \item Fig.~\ref{fig:relevant} shows the ending states of the timelines corresponding to \textit{Session 1} and \textit{Session 2} in Fig.~\ref{fig:patterns}. The model labels both sessions as sessions in which search was relevant to the purchase, but the geometric interpretation of the underlying timelines highlights the big difference between the two cases: in the ``lakers'' case, search is relevant but the shopper was already interested in that part of the space -- in other words, she was likely to found that product anyway; in the ``boardshorts'' case, the original (\textit{orange}) \textit{and} the \textit{after search} timelines (\textit{blue}) are very close, while the \textit{no search} (\textit{green}) timeline ends up in a very different region, since that future diverges quickly as the model predominantly explore the sneaker portion of the space. In this case, search influence on the final conversion is larger.
    \item Fig.~\ref{fig:irrelevant} shows the ending states of the timelines corresponding to \textit{Session 3} in Fig.~\ref{fig:patterns} plus a multi-intent, not related session (not shown for brevity). The model correctly labels both sessions as sessions in which search was \textit{not} relevant to the purchase -- also in this case, the geometric interpretation provides an intuitive understanding of the model behavior: the three timelines in both cases are very far from each other in the product space, with no immediate correlation between the target points (compare with Fig.~\ref{fig:relevant}).
\end{enumerate}

\section{Generating a Synthetic Dataset}
\label{app:synthetic}
Generating synthetic datasets with known dependencies is a common strategy in the causal literature, since labels on real-world datasets are impossible to obtain. We detail here the process of generating a synthetic dataset starting from an existing \textit{Shop S}, by dividing the final dataset in four major subsets: non-converting sessions $NC$, converting session without search $CW$, converting sessions related to a search query $CS$, converting session unrelated to search interaction $CU$. To capture the qualitative subtleties analyzed in Section~\ref{sec:methodology}, $CS$ sessions fall in turn in three buckets: single-intent, ``exact match'' queries $SE$; multi-intent, ``exact match'' queries $ME$; single-intent, ``related match'' queries $SR$. The parameters $k$ and $r$ control the cardinality $C(D)$ of the resulting synthetic dataset $D$, as specified below:

\begin{itemize}
    \item $C(CW)=k$ and $C(NC)=k * 2$;
    \item $C(CU)=k$;
    \item $C(SE)=k * r$ where $0.0 \leq r \leq 1.0$;
    \item $C(ME)=k * r / 2$;
    \item $C(SR)=k * r / 4$.
\end{itemize}

We generate $NC$, $CS$ and $CU$ in the planned cardinality with these methods:

\begin{itemize}
    \item \textbf{NC}: we sample $C(NC)$ non-converting sessions from a real \textit{Shop S};
    \item \textbf{CU}: we sample $C(CU)$ converting sessions \textit{without} search interactions; we inject in each session a search interaction randomly sampled from the dataset; the insertion position is sampled proportionally to empirical frequencies in converting sessions with search queries;
    \item \textbf{SE}: we sample $C(SE)$ non-converting sessions with search interaction involving a click on product $Px$; sampling from the underlying shop space in proportion to vector similarity, we generate up to $e=3$ random product interactions between $Px_{click}$ and a synthetic event $Px_{add}$, and do the same for $i=2$ simulated actions between $Px_{add}$ an $Px_{purchase}$.
    \item \textbf{ME}: we sample a random starting product $Px$ uniformly from the product space and generate up to $y=4$ product interaction repeatedly sample new items in proportion to vector similarity. We then sample $n=20$ search interactions from search sessions, and select the query+click pair which is the farthest from the current product (to simulate a ``jump'' in the product space as in Fig.~\textbf{\ref{fig:patterns}.2}). After the click, we proceed as for \textbf{SE} above: we generate up to $e=3$ random interactions between $Px_{click}$ and a synthetic event $Px_{add}$, and do the same for $i=2$ simulated actions between $Px_{add}$ an $Px_{purchase}$
    \item \textbf{SR}: we sample $C(SR)$ non-converting sessions  with search interaction involving a click on product $Px$; we use the underlying space to select a target purchase of nearby product $Rx$; we then proceed to generate up to $e=3$ random interactions between $Px_{click}$ and a synthetic event $Rx_{add}$, and do the same for $i=2$ simulated actions between $Rx_{add}$ an $Rx_{purchase}$.
\end{itemize}

$i$, $e$ and $y$ above are chosen after some empirical tests in order to produce shopping sequences which represent realistic browsing for shoppers while at the same time maintain a uniform ``underlying intent'' to test the model ability in picking-up the relevant long-range behavioral dependencies.

It is important to note that the methodology is pretty general, insofar as it relies on \textit{prod2vec} as a proxy for semantic product proximity: if more/different qualitative patterns of causal influence are needed, it is straightforward to extend the generating procedure to include them as well.

\end{document}